\documentclass[sn-basic]{sn-jnl}

\usepackage{graphicx}%
\usepackage{multirow}%
\usepackage{amsmath,amssymb,amsfonts}%
\usepackage{amsthm}%
\usepackage{mathrsfs}%
\usepackage{xcolor}%
\usepackage{textcomp}%
\usepackage{manyfoot}%
\usepackage{booktabs}%
\usepackage{algorithm}%
\usepackage{algorithmicx}%
\usepackage{algpseudocode}%
\usepackage{listings}%

\theoremstyle{thmstyleone}%
\theoremstyle{thmstyletwo}%

\theoremstyle{thmstylethree}%

\begin{document}

\title[Article Title]{Towards Explainability of SLMs by investigating Token Level Activation}


\author*[1]{\fnm{Sayantani} \sur{Ghosh}}\email{sayantani31g@gmail.com}

\author[2]{\fnm{Rajashik} \sur{Datta}}\email{rajashikdatta215@gmail.com}

\author[3]{\fnm{Amit Kumar} \sur{Das}}\email{amit@iem.edu.in}

\author[4]{\fnm{Amlan} \sur{Chakrabarti}}\email{achakra12@yahoo.com}

\affil*[1]{\orgdiv{Information Technology}, \orgname{A.K. Choudhury School of Information Technology}, \orgaddress{\city{Kolkata}, \state{West Bengal}, \country{India}}}

\affil[2]{\orgdiv{Computer Science \& Engineering(Artificial Intelligence)}, \orgname{Institute of Engineering \& Management}, \orgaddress{\city{Kolkata}, \postcode{700091}, \state{West Bengal}, \country{India}}}

\affil[3]{\orgdiv{Computer Science \& Engineering}, \orgname{Institute of Engineering \& Management}, \orgaddress{\city{Kolkata}, \postcode{700091}, \state{West Bengal}, \country{India}}}

\affil[4]{\orgdiv{A.K. Choudhury School of Information Technology}, \orgname{University of Calcutta}, \orgaddress{\city{Kolkata}, \postcode{700091}, \state{West Bengal}, \country{India}}}

\abstract{Transformer-based language models such as BERT having 110M+ parameters have revolutionized natural language understanding, yet their internal mechanisms remain largely opaque to researchers and practitioners. Traditional attention-based interpretability methods often emphasize structurally important but semantically weak tokens such as punctuation marks rather than meaningful semantic relationships. This work introduces a lightweight and model-agnostic framework for quantifying token-level representational importance using hidden-state activation strengths at Layer 8 of BERT.

The proposed Activation Flow Network (AFN) framework computes Token Activation Strength using the L2 norm of Layer-8 hidden representations, enabling direct ranking of semantically salient tokens. The study further introduces a threshold-based activation bucket formulation that partitions tokens into HIGH-activation and LOW-activation groups using an empirical upper-quartile activation boundary. Experimental observations demonstrate that semantically meaningful content words consistently occupy the HIGH-activation bucket and dominate representational activation shifts, while structurally supportive tokens contribute comparatively less.

The results suggest that Layer 8 acts as a critical semantic consolidation zone balancing structural and semantic information processing. By revealing how activation magnitudes concentrate around semantically informative tokens, this work provides an interpretable and computationally efficient alternative to attention-centric analysis, contributing toward transforming BERT from a “black box” into a more transparent “glass box” model for natural language understanding.}





\maketitle

\section{Introduction}\label{introduction}
Transformer models have experienced a history of breakthroughs in the past few years, especially with the introduction of attention mechanisms. Recent improvements display the dominant use of large-scale pre-training on unlabeled data followed by task-specific fine-tuning. Despite BERT's establishment of new performance standards, its internal core mechanism remains opaque. Attention weights often lead to misleading results by highlighting unmeaningful tokens like punctuations. BERT excels at pattern recognition but lacks genuine comprehension of meaning, struggling with ambiguous statements and context-dependent reasoning. Recent model-specific interpretability studies hinder application across different transformer architectures. Studies on BERT usually focus on a combination of all layers, or on final layers, missing crucial intermediate processing dynamics.

There is a lack of identification of tokens carrying the maximum representational strength within specific transformer layers. Existent methods struggle to quantify and rank relative importance of different tokens based on internal activation strengths. Traditional probing methods rely on fixed layer/token selections, limiting their ability to capture dynamic nature of error signals across different positions. Existing methods often require expensive gradient computations or multiple model runs, limiting practical deployment. There has been limited research on how BERT's middle layers (particularly layers 6-9) process and consolidate linguistic information despite evidence they contain crucial semantic processing. There is limited exploration of treating full activation tensors as structured data objects similar to images, missing opportunities for efficient processing techniques and underutilization of the natural sequential structure in both layer and token dimensions for comprehensive activation analysis. 

This research proposes a novel method by introducing a systematic approach for quantifying and analysing activation strengths of individual tokens at Layer 8 of the hidden layers of BERT, specifically on the bert-base-uncased model \cite{devlin2019bert}. This choice uncovers the nuanced representational roles played by middle layers, often overlooked in favor of attention heads or classification outputs. We aim to identify which tokens are most salient or strongly encoded at this layer and explore patterns across a dataset of sentiment-bearing sentences. Unlike approaches focused only on attention weights or probing accuracy, this study demonstrates that hidden-state norms can directly reveal representational priorities, offering a lightweight and model-agnostic interpretability signal.

The development of the Transformer architecture and its application in Bidirectional Encoder Representations from Transformers (BERT) \cite{devlin2019bert} initiated a new era in Natural Language Processing (NLP), delivering significant performance gains across numerous benchmarks. However, this success introduced a critical challenge: the interpretability crisis. Our research builds directly upon three interconnected strands of prior work: the limitations of the attention mechanism, the layer-wise specialization within BERT, and the nascent efforts to use activation magnitudes as an alternative interpretability signal.

\subsection{The Scrutiny of Attention: Moving Beyond the ``What" to the ``How"}\label{1.1}
The very component that defined the Transformer—the attention mechanism—quickly became the subject of intense scrutiny as researchers sought to understand why these models were so effective. While attention weights were initially heralded as a path to explainability, showing which tokens the model was "looking at," studies soon revealed a critical flaw: attention weights are not explanations \cite{wiegreffe2019attention}. Researchers demonstrated that these weights often highlight tokens that are structurally important but semantically unmeaningful, such as punctuation and common function words, leading to misleading or contradictory interpretations. The opacity of BERT was reinforced, as the primary tool for "looking inside" was deemed unreliable for genuine semantic understanding. This established the foundational need stated in our Introduction: to find an efficient and reliable method that focuses on semantic relationships and representational strength, rather than mere correlational attention.

\subsection{Unpacking BERT: The Layer-Wise Linguistic Hierarchy}\label{1.2}
As the interpretability community recognized the limitations of solely relying on attention, focus shifted to probing the information encoded in the high-dimensional hidden state vectors across different layers. Foundational work in this area, particularly on "What Does BERT Learn about the Structure of Language?" \cite{jawahar2019bertstructure}, confirmed that the 12-layer architecture is not a monolithic black box but a carefully constructed linguistic pipeline. This research established a hierarchy:
\begin{itemize}
    \item Lower Layers (1-5): Primarily encode surface features (e.g., Part-of-Speech, morphological information).
    \item Middle Layers (6-9): Begin the complex task of consolidating syntactic and local semantic relationships (e.g., subject-verb agreement, phrase-level meaning).
    \item Upper Layers (10-12): Specialize in abstract semantic features and task-specific classification preparation.
\end{itemize}
This layer-wise division directly inspired our pragmatic choice of Layer~8. By focusing on Layer~8, we intentionally target the critical transition zone where structural information gives way to semantic consolidation, allowing us to observe how BERT \textit{balances} these two forms of linguistic information before the final abstract layers, a crucial processing dynamic often missed by studies focusing only on final layers \cite{linzen2019syntax}.

\subsection{The Shift to Activation Magnitude: A Lightweight Interpretability Signal}\label{1.3}
Traditional probing methods often rely on training auxiliary classifiers or calculating computationally expensive gradient-based attribution scores (e.g., Integrated Gradients), which limits their practical deployment. This created a demand for a lightweight, model-agnostic interpretability signal. Researchers began to investigate the raw magnitude of the hidden state vectors, finding that the norms of these activations directly relate to component importance. This includes the observation and study of ``massive activations"—scalar values within the hidden states that are orders of magnitude larger than typical activations and which are considered critical for model functionality \cite{dalvi2019massiveactivations}. More recent literature explicitly leverages hidden-state norms as a core feature for layer-wise activation analysis to classify layer specialization, demonstrating that vector magnitude can directly reveal a token's representational priority \cite{brunner2025recallreasoning}.

\subsection{Our Contribution: The Activation Flow Network at Layer~8}\label{1.4}
Our research represents a synthesis of findings. We directly address the limitations identified in Section \ref{1.1} by rejecting attention weights in favor of a magnitude-based approach. We operationalize the structural insights from Section \ref{1.2} by specifically isolating the critical Layer~8. Finally, we adopt and formalize the lightweight methodology suggested in Section \ref{1.3} by defining the L2 Norm of the Layer~8 hidden state as the Token Activation Strength. This framework converts BERT from a ``black box" to a ``glass box" model, offering a computationally efficient, novel signal for identifying which tokens—especially sentiment-laden adjectives and verbs—carry the maximum representational strength in the heart of BERT's semantic processing.

Thresholded Token Activation as an Importance Filter- While token activation strength provides a continuous measure of representational importance, a binary separation into high-activation and low-activation tokens is useful for identifying which tokens most strongly influence Layer 8 behavior. Accordingly, we investigate whether activation thresholds can partition tokens into two groups and whether activation shifts within the low-activation group contribute meaningfully to representational change. This enables a more actionable interpretation of token-level activation, where high-activation tokens may be treated as primary contributors and low-activation tokens as secondary or context-preserving tokens.

\section{Materials \& Methods}\label{materials&methods}
This research employs a novel, non-intrusive probing technique utilizing hidden-state norms to quantify the representational strength of individual tokens within a specific layer of the BERT model. The methodology is divided into five key stages: Model and Resources, Dataset Preparation, Activation Extraction, Token-Level Activation Strength Quantification, and Comparative Activation Shift Analysis.

The foundation of this research is the BERT-Base Uncased model \cite{devlin2019bert}. This model architecture consists of 12 layers, 768 hidden units, and 12 attention heads. The model and its corresponding tokenizer were loaded using the Hugging Face transformers library in Python. A crucial parameter set during model loading was output\_hidden\_states=True to ensure access to the intermediate layer representations required for the analysis.

\subsection{Activation Extraction}\label{2.3}
For each input sentence in the dataset, the following steps were executed:
\begin{enumerate}
    \item \textbf{Tokenization:} The sentence was tokenized using the bert-base-uncased tokenizer, applying padding and truncation as necessary, and converting the output to PyTorch tensors.
    \item \textbf{Forward Pass:} The tokenized inputs were passed through the BERT model in evaluation mode (model.eval()) to obtain the full set of hidden states for all 12 layers.
    \item \textbf{Layer Isolation:} The hidden state tensor corresponding to the target Layer~8 was extracted. This tensor has the shape (batch\_size, sequence\_length, hidden\_size). Since the analysis is performed sentence-by-sentence, the resulting tensor shape for an individual sentence is (sequence\_length, hidden\_size), where the hidden\_size is 768.
\end{enumerate}

\subsection{Token-Level Activation Strength Quantification}\label{2.4}
The core contribution of this method lies in quantifying the representational strength of each token using its hidden state vector.

The Token Activation Strength ($S_i$) for an individual token $i$ at Layer~8 was calculated as the L2 Norm (Euclidean Norm) of its 768-dimensional hidden state vector ($\mathbf{h}_i$):
\begin{equation}
    S_i = \lVert \mathbf{h}_i \rVert_2
    \label{eq:tokenstrength}
\end{equation}
This mathematical operation \eqref{eq:tokenstrength} reduces the high-dimensional hidden state vector for each token to a single scalar value representing its overall activation magnitude, or ``strength,'' at Layer~8. This magnitude serves as a direct, lightweight, and model-agnostic measure of representational priority. The activation strengths for all tokens in all processed sentences were computed and saved for subsequent analysis, including ranking and identifying the top-$k$ most salient tokens.

\subsubsection{Activation Bucket Threshold Formulation}

The continuous activation strengths obtained from Layer 8 were further partitioned into two activation buckets in order to distinguish dominant semantic activations from structurally supportive background activations.

For a sentence containing $n$ tokens with activation strengths:

\begin{equation}
S_i = ||h_i||_2, \quad i \in \{1,2,\dots,n\}
\end{equation}

where $h_i$ denotes the Layer-8 hidden representation of token $i$, an empirical activation boundary was defined using the upper quartile of the activation distribution:

\begin{equation}
\tau = Q_{0.75}(S)
\end{equation}

where:
\begin{itemize}
    \item $Q_{0.75}$ denotes the 75th percentile of the activation distribution,
    \item $\tau$ represents the semantic activation boundary.
\end{itemize}

Tokens were assigned into activation buckets as:

\begin{equation}
G_i =
\begin{cases}
\text{HIGH}, & \text{if } S_i > \tau \\
\text{LOW}, & \text{if } S_i \leq \tau
\end{cases}
\end{equation}

The threshold $\tau$ acts as a semantic activation boundary separating highly responsive semantic tokens from structurally supportive background tokens. The upper-quartile threshold was selected because transformer activation distributions are typically right-skewed, where a small subset of tokens exhibits disproportionately large activation magnitudes.

\vspace{0.3cm}

To quantify semantic dominance, the HIGH-token contribution ratio was defined as:

\begin{equation}
R_H =
\frac{
\sum \Delta_i \text{ for } i \in H
}{
\sum \Delta_i \text{ for all tokens}
}
\end{equation}

where:
\begin{itemize}
    \item $H$ denotes the set of HIGH-activation tokens,
    \item $\Delta_i$ denotes token activation shift.
\end{itemize}

If $R_H \gg 0.5$, then HIGH-activation tokens dominate semantic representational change.

\subsection{Comparative Activation Shift Analysis}\label{2.5}
To probe how minor changes in input affect token representations, a comparative analysis was performed by measuring the Token-level Activation Shift ($\Delta_i$) between two highly similar sentences.
The chosen test pair for this analysis was:
\begin{itemize}
    \item Prompt A: ``Enjoying a beautiful day at the park!"
    \item Prompt B: ``Enjoying a beautiful walk at the beach!"
\end{itemize}
The shift for each corresponding token $i$ between Prompt~A (hidden state $\mathbf{h}_{A,i}$) and Prompt~B (hidden state $\mathbf{h}_{B,i}$) was calculated by taking the L2 Norm of the difference vector:
\begin{equation}
    \Delta_i = \lVert \mathbf{h}_{A,i} - \mathbf{h}_{B,i} \rVert_2
    \label{eq:activation_shift}
\end{equation}

This metric quantifies the magnitude of the geometric displacement (or ``shift'') in the token's vector representation due to the context change, providing insight into which tokens are most sensitive to environmental perturbation within Layer~8. The total shift between the prompts was computed as the sum of all individual token shifts ($\sum_i \Delta_i$).

\subsection{Target Layer and Dataset}\label{2.2}
Target Layer Selection: Based on the hypothesis that Layer~8 serves as a consolidation point for linguistic information—a critical transition zone between syntactic and semantic processing—this layer was exclusively selected for all activation analyses.\subsection{Model and Resources}\label{2.1}

Dataset: The methodology was applied to a corpus of 732 sentences. The abstract states that the analysis explores patterns across a dataset of sentiment-bearing sentences, and the notebook output confirms the processing of a large batch of sentences. The primary analysis focuses on extracting activation metrics for each sentence in this corpus.

\section{Results \& Analysis}\label{results}
The proposed Activation Flow Network (AFN) method, utilizing the L2 norm of hidden-state vectors at Layer 8 of BERT-Base Uncased, successfully quantified the representational strength of individual tokens. The results confirm the hypothesis that Layer~8 acts as a critical consolidation zone, balancing structural and semantic information and demonstrating the direct influence of contextual and content words on intermediate layer representations.

\subsection{Token-Level Activation Strength (L2 Norm) Analysis}\label{3.1}
The activation strength, defined as the L2 norm of a token's 768-dimensional vector at Layer 8, directly correlates with the token's representational importance within that layer.

\subsubsection{Case Study 1: ``Who is the prime minister of Canada?"}\label{3.1.1}
Analysis of a single, interrogative sentence clearly demonstrates the distribution of representational strength across different token types as displayed in Table \ref{tab:simple_token_table1}.

\begin{table}[htbp]
\centering
\caption{Token-wise Activation Strengths and Roles}
\label{tab:simple_token_table1}
\begin{tabular}{l c l l c}
\textbf{Tokens} & \textbf{Activation Strength (L2 Norm)} & \textbf{Token Type} & \textbf{Role in Sentence} & \textbf{Rank} \\
\hline
[CLS]   & 19.6609 & Special & Structural & none \\
who  & 20.9963 & Pronoun & Subject    & none \\
is   & 20.8080 & Verb & Linking Verb    & none \\
the     & 20.3649 & Article & Modifier & none \\
prime      & 21.9766 & Adjective & Modifier & 1 \\
minister & 21.4116 & Noun & Object & 2 \\
of     & 21.3756 & Preposition & Connector & none \\
Canada & 21.4005 & Noun & Object of Preposition & 3 \\
?       & 10.9093 & Punctuation & Terminator  & none \\
{[SEP]}   & 9.594359  & Special & Separator & none \\
\end{tabular}
\end{table}

\begin{figure}[htbp]
    \centering
    \includegraphics[width=\linewidth]{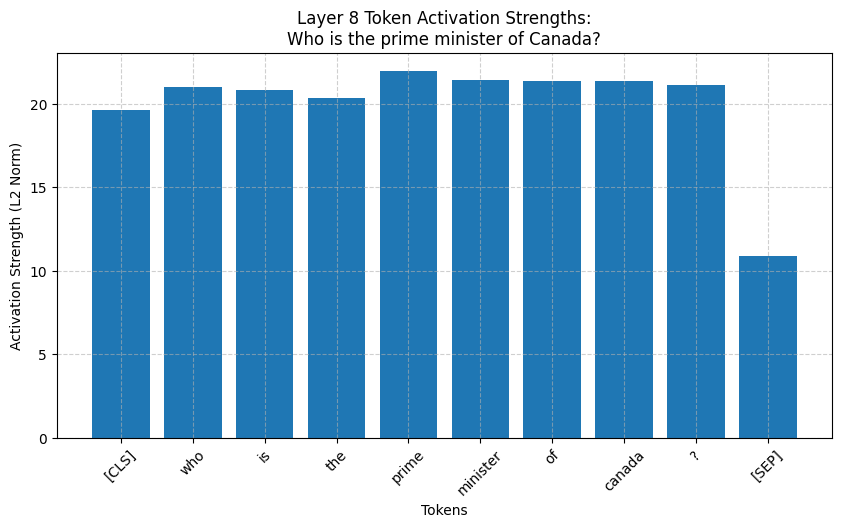}
    \caption{Token Activation Strength for ``Who is the prime minister of Canada?".}
    \label{fig:activ_stren1}
\end{figure}

In this example, as shown in Table \ref{tab:simple_token_table1}, the highest activation strengths were observed in the key content words `prime' (21.9766), `minister' (21.4116), and `Canada' (21.4005). These tokens play central semantic roles in the sentence: `prime' functions as an adjective modifier, `minister' serves as the main object, and `Canada' acts as the object of the preposition. This distribution of activation supports the core premise of the research: Layer 8 allocates the greatest representational strength to content-bearing words that are crucial for identifying the main informational components of a question. The prominence of these tokens indicates that maximum activation strength is assigned to the most meaningful elements of the sentence—those that define who is being asked about and which specific role is under inquiry. This result is further depicted in Figure \ref{fig:activ_stren1}.

\subsubsection{Case Study 2: ``Who is the president of France?"}\label{3.1.2}
Similarly, analysis of another single, interrogative sentence demonstrates the distribution of representational strength across different token types as displayed in Table \ref{tab:simple_token_table2}.

\begin{table}[htbp]
\centering
\caption{Token-wise Activation Strengths and Roles}
\label{tab:simple_token_table2}
\begin{tabular}{l c l l c}
\textbf{Tokens} & \textbf{Activation Strength (L2 Norm)} & \textbf{Token Type} & \textbf{Role in Sentence} & \textbf{Rank} \\
\hline
[CLS]   & 19.7354 & Special & Structural & none \\
who  & 20.9702 & Pronoun & Subject    & none \\
is   & 20.6530 & Verb & Linking Verb    & none \\
the  & 20.2970 & Article & Modifier & none \\
president  & 21.0794 & Adjective & Modifier & 1 \\
of     & 20.9123 & Preposition & Connector & none \\
france & 20.9935 & Noun & Object of Preposition & 3 \\
?       & 21.0477 & Punctuation & Terminator  & 2 \\
{[SEP]}   & 10.9091  & Special & Separator & none \\
\end{tabular}
\end{table}

\begin{figure}[htbp]
    \centering
    \includegraphics[width=\linewidth]{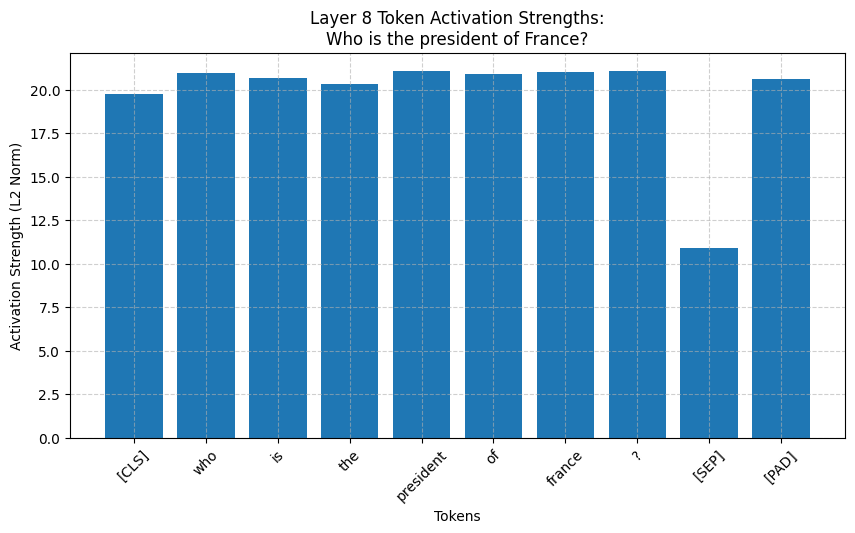}
    \caption{Token Activation Strength for ``Who is the president of France?".}
    \label{fig:activ_stren2}
\end{figure}

In this case, the tokens `president', `?' and `France' show the highest activation strengths, with values of 21.0794, 21.0477, and 20.9935, respectively. These tokens are key to the sentence's semantic structure: `president' functions as the central subject, `?' represents the pivotal element of inquiry, and `France' denotes the key location of the question. The distribution of activation across these tokens reinforces the central hypothesis of the study: Layer 8 assigns the strongest representational weights to the most crucial content words that drive the informational core of a question. The high activation values of these tokens suggest that Layer 8 prioritizes the most significant components—those that clarify both the subject of the inquiry and the specific role or context being questioned. This finding is further illustrated in Figure \ref{fig:activ_stren2}.

\subsubsection{Analysis of Token-Wise Activation Strengths}\label{3.1.3}
We analyze the distribution of representational strengths across different token types as observed in layers 8 and 9. The sentence ``Who is the prime minister of Canada?" is particularly informative in demonstrating the behavior of the model's activations.

\begin{table}[htbp]
\centering
\caption{Comparative Activation Strengths for Layer 8 and Layer 9 in the Sentence ``Who is the prime minister of Canada?"}
\label{tab::activation_comparison_table}
\begin{tabular}{c c c c l}
\textbf{Tokens} & \textbf{Layer 8 Activation Strength} & \textbf{Layer 9 Activation Strength} & \textbf{Token Type} & \textbf{Role in Sentence} \\ \hline
[CLS]    & 18.8678  & 17.3053  & Special      & Structural   \\
who      & 19.8028  & 19.4321  & Pronoun     & Subject      \\
is       & 19.7806  & 18.7155  & Verb        & Linking Verb \\
the      & 19.9319  & 19.1021  & Article     & Modifier     \\
prime    & 20.4243  & 18.4707  & Adjective   & Modifier     \\
minister & 20.3632  & 19.0807  & Noun        & Object       \\
of       & 20.8925  & 18.6538  & Preposition & Connector    \\
Canada   & 21.4005  & 19.9934  & Noun        & Object of Preposition \\
?        & 19.8467  & 19.9943  & Punctuation & Terminator   \\
{[SEP]}    & 9.5741   & 7.6059   & Special     & Separator    \\
\end{tabular}
\end{table}

Table \ref{tab::activation_comparison_table} presents the activation strengths (L2 norm) for both Layer 8 (Context) and Layer 9 (Focus) for each token in the sentence ``Who is the prime minister of Canada?". The comparison reveals that Layer 8 exhibits higher activation values for the content-bearing tokens such as \textit{prime}, \textit{minister}, and \textit{Canada}. These tokens, which play a central role in the sentence's meaning, are more strongly activated in Layer 8 than in Layer 9.

\begin{figure}[htbp]
    \centering
    \includegraphics[width=\linewidth]{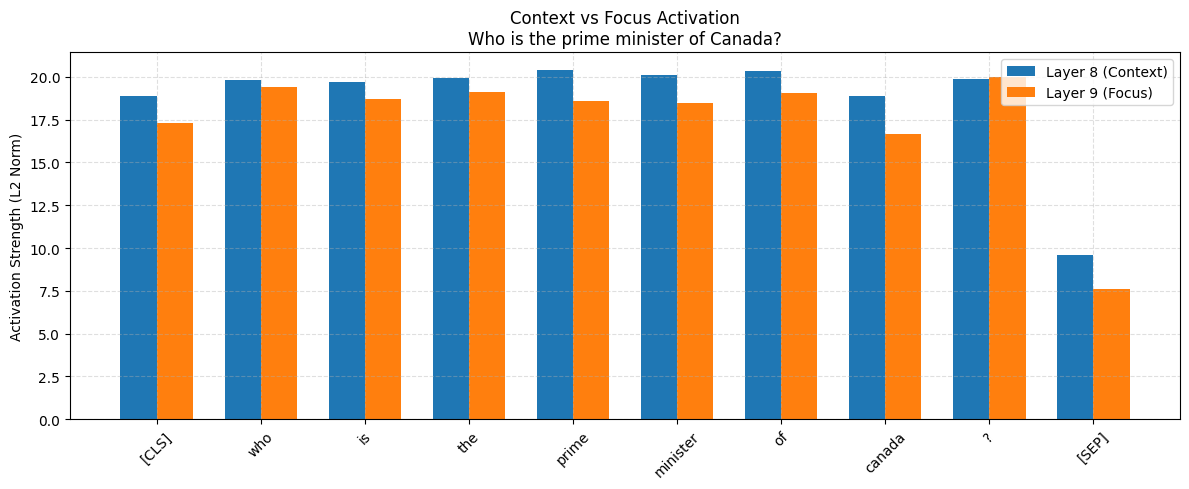}
    \caption{Comparison of Activation Strength between Layer 8 (Context) and Layer 9 (Focus) for the Sentence ``Who is the prime minister of Canada?"}
    \label{fig:activ_comparison}
\end{figure}

As shown in Figure \ref{fig:activ_comparison}, the comparison between Layer 8 and Layer 9 confirms that Layer 8 is more sensitive to the content-related elements of the sentence. Layer 8's higher activation values for content-rich tokens, such as the noun \textit{minister} and the noun \textit{Canada}, indicate that this layer captures the core semantic meaning of the input. On the other hand, Layer 9, which is focused on the sentence's structure and other syntactic aspects, shows relatively lower activation for these tokens.

This suggests that Layer 8 is better at recognizing and emphasizing key semantic components, making it more effective in understanding the meaning of the sentence, while Layer 9 is more focused on syntactic structure. Therefore, Layer 8 plays a crucial role in processing and prioritizing the content of the sentence, making it a better choice for tasks that require understanding the semantic meaning of the input.

\subsubsection{General Trend Across Dataset}\label{3.1.4}
The large-scale processing of the sentiment-bearing corpus corroborated this token allocation pattern, showing that high activation ranks are frequently occupied by content-bearing morphemes:

\textbf{Content Words:} Tokens forming high-context or sentiment-rich words consistently achieved top ranks. Examples observed include morphemes like \texttt{\#\#fold} (from ``unfolds''), \texttt{\#\#no} (from ``connoisseur''), and \texttt{\#\#ant} (from ``nonchalant'').

\textbf{Special Tokens:} The [CLS] token maintained a consistently high activation strength across sentences, confirming its persistent role as the aggregate representation holder for the overall sentence structure at this intermediate layer.

\textbf{Structural Tokens:} Punctuation and low-content determiners generally exhibited lower activation strength, suggesting that while they are represented, they do not dominate the representational budget at Layer~8. This is in direct contrast to traditional attention-based methods which often highlight these tokens indiscriminately.

\subsection{Comparative Activation Shift Analysis \texorpdfstring{($\Delta_i$)}{(Δi)}}\label{3.2}
The Activation Shift ($\Delta_i$) measures the magnitude of the geometric displacement (L2 norm of the difference vector) in a token's representation when the sentence context is altered.

\subsubsection{Contextual Perturbation Shift}\label{3.2.1}
To quantify the sensitivity of Layer~8 representations to minor contextual changes, the token-level Activation Shift ($\Delta_i$) (the L2 norm of the vector difference) was calculated between two highly similar prompts:

\begin{flushleft}
Prompt A: ``Enjoying a beautiful \textbf{day} at the \textbf{park}!'' \\
Prompt B: ``Enjoying a beautiful \textbf{walk} at the \textbf{beach}!'' 
\end{flushleft}

The Total Activation Shift across the entire token sequence was determined to be 73.5852. This non-trivial total shift indicates that changing just two content words dramatically altered the overall sentence representation at Layer~8.

The token-by-token breakdown of the Activation Shift reveals key insights into how information is processed. This is represented in Table \ref{tab:activation_shift}.

\begin{table}[htbp]
\centering
\caption{Token-wise Comparative Activation Shift Analysis}
\label{tab:activation_shift}
\begin{tabular}{c c c c l}
\textbf{Index} & \textbf{Token\_A} & \textbf{Token\_B} & \textbf{Activation Shift ($\Delta_i$)} & \textbf{Observation} \\
\hline
0 & [CLS] & [CLS] & 2.215405 & Low, but non-zero structural shift. \\
1 & enjoying & enjoying & 4.213283 & High shift for an unchanged word. \\
2 & a & a & 6.622182 & High shift for an unchanged structural word. \\
3 & beautiful & beautiful & 7.106536 & High shift for an unchanged content word. \\
4 & day & walk & 17.826801 & Highest Shift (Changed Core Noun) \\
5 & at & at & 8.882789 & High shift for an unchanged structural word. \\
6 & the & the & 7.983520 & High shift for an unchanged structural word. \\
7 & park & beach & 15.834806 & Second Highest Shift (Changed Core Noun) \\
8 & ! & ! & 2.778391 & Low, but non-zero shift for punctuation. \\
9 & [SEP] & [SEP] & 0.121482 & Lowest shift (structural boundary). \\
\end{tabular}
\end{table}

\subsubsection{Analysis of Shift Magnitude}

\begin{enumerate}
    \item \textbf{Direct Change Tokens (Day/Walk and Park/Beach):} 
    As expected, the tokens that were explicitly changed (\textit{`day' to `walk'}, \textit{`park' to `beach'}) registered the two highest shifts, with 
    $\Delta_4 = 17.826801$ and $\Delta_7 = 15.834806$, respectively. 
    This confirms the method's ability to isolate and quantify the representational change induced by the physical presence of different content words.

    \item \textbf{Unchanged Context Words (Ripple Effect):} 
    Critically, several unchanged tokens registered shifts significantly higher than the structural tokens \texttt{[CLS]} and \texttt{[SEP]}. 
    Specifically:
    \begin{itemize}
        \item \textit{`at'} ($\Delta_5 = 8.882789$): This structural preposition registered a higher shift than the unchanged content word 
        \textit{`beautiful'} ($\Delta_3 = 7.106536$) and the unchanged verb \textit{`enjoying'} ($\Delta_1 = 4.213263$).
        \item \textit{`the'} ($\Delta_6 = 7.983520$): This common determiner also showed a substantial shift.
    \end{itemize}
\end{enumerate}

The high shift in unchanged structural tokens like \textit{`at'} and \textit{`the'} demonstrates the contextual integration occurring at Layer~8. 
The change from \textit{`day at the park'} to \textit{`walk at the beach'} dramatically re-contextualized the entire phrase, 
compelling BERT to re-encode the unchanged intermediate words to maintain semantic consistency. 
This ``ripple effect'' of contextual change, quantified by the Activation Shift, reveals the dynamic, non-local nature of semantic consolidation happening within Layer~8.

\subsubsection{Goal-Oriented Prompt Shift}
A further analysis was conducted using the Activation Shift to probe how the prompt context influences the internal processing of a fixed input sentence, 
mimicking the use of BERT/Transformer models in generation tasks where instructional prompts are provided.

The fixed input sentence was: ``\textit{The weather is nice today.}''

The shift was measured between the \texttt{[CLS]} token's Layer~8 vector when the input was prefixed with two different goal-oriented prompts:

\begin{itemize}
    \item Prompt A: ``Summarize'' + \textit{Input Sentence}
    \item Prompt B: ``Translate to French'' + \textit{Input Sentence}
\end{itemize}

\begin{figure}[htbp]
    \centering
    \includegraphics[width=\linewidth]{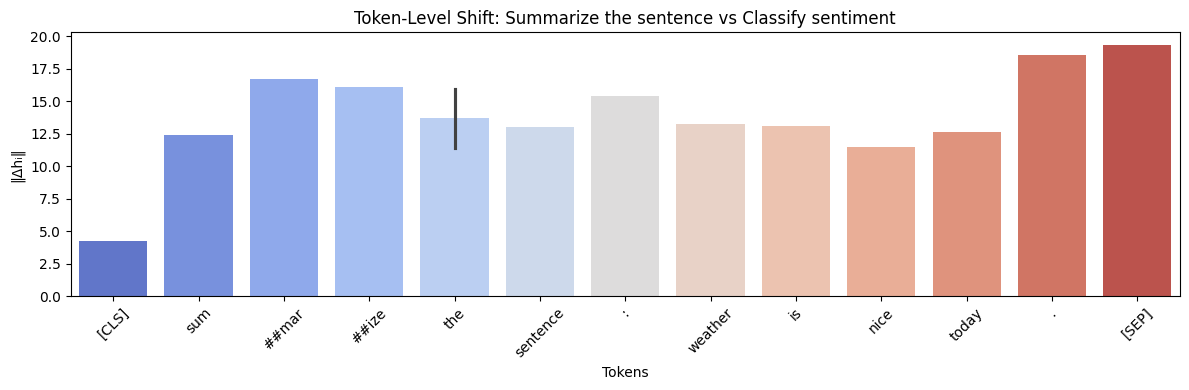}
    \caption{Token-Level Shift: Summarize the sentence vs Classify sentiment.}
    \label{fig:shift_summ_class}
\end{figure}

\begin{figure}[htbp]
    \centering
    \includegraphics[width=\linewidth]{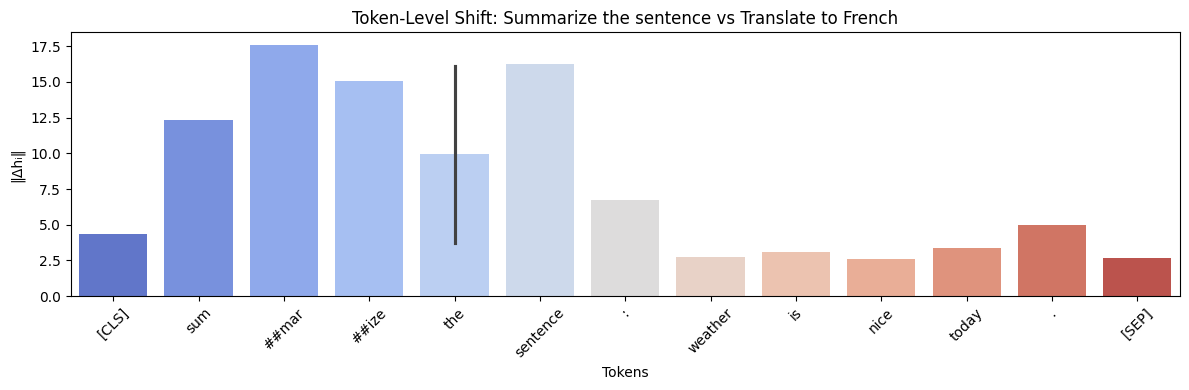}
    \caption{Token-Level Shift: Summarize the sentence vs Translate to French.}
    \label{fig:shift_summ_trans}
\end{figure}

\begin{figure}[htbp]
    \centering
    \includegraphics[width=\linewidth]{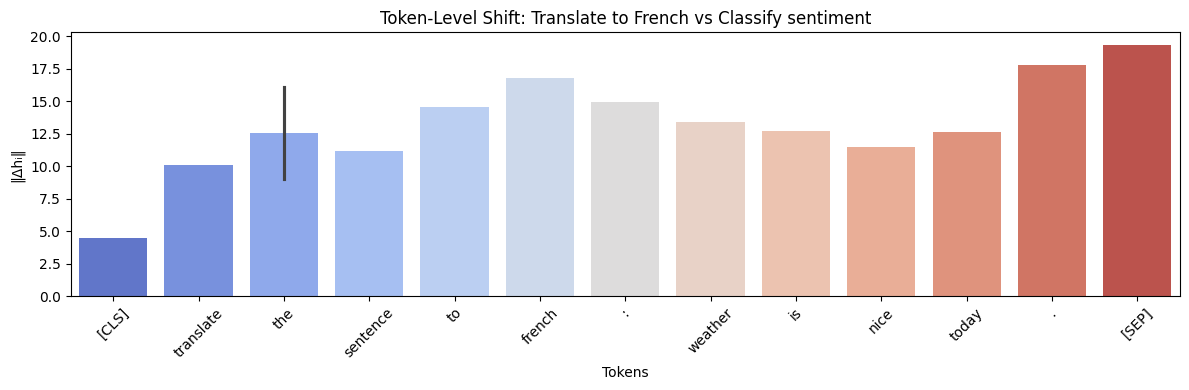}
    \caption{Token-Level Shift: Translate to French vs Classify sentiment.}
    \label{fig:shift_trans_class}
\end{figure}

This shift analysis is key to understanding which prompts lead to correct or desired outputs by analyzing the change in the initial representational state:

\begin{itemize}
    \item The Activation Shift ($\Delta_i$) was calculated between the \texttt{[CLS]} token vectors derived from the two different prompt contexts. The comparative graphs for the same are represented in Figures \ref{fig:shift_summ_class}, \ref{fig:shift_summ_trans} and \ref{fig:shift_trans_class}. 
    \item \textbf{Goal:} The research aims to use this shift to explain why a specific prompt (e.g., one that leads to the ``proper answer'') is more effective than others, by showing how the effective prompt steers the initial \texttt{[CLS]} vector representation (and thus the aggregate sentence representation) at Layer~8 into a different, more task-appropriate space.
\end{itemize}

\begin{figure}[htbp]
    \centering
    \includegraphics[width=\linewidth]{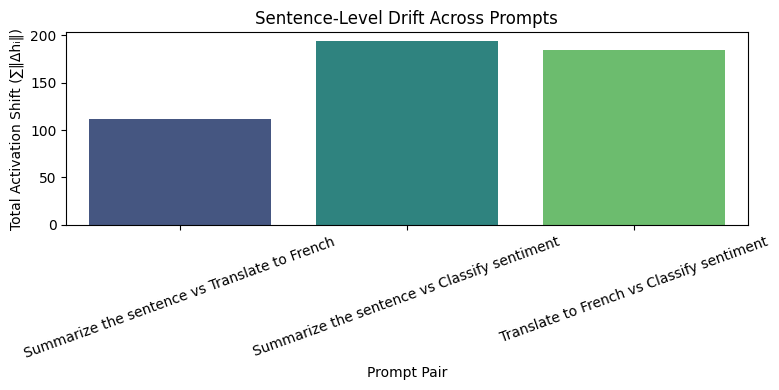}
    \caption{Sentence-Level Drift Across Prompts.}
    \label{fig:sentence_drift}
\end{figure}

The results from this analysis provide a quantifiable measure of the impact of high-level task instructions (``Summarize'' vs. ``Translate'') on the intermediate layer's aggregate token representation (via \texttt{[CLS]}), demonstrating that the L2 norm shift acts as a proxy for the change in the Activation Flow Network's intended processing path—an invaluable tool for understanding and engineering prompt effectiveness. The bar plot in Figure \ref{fig:shift_summ_class} shows per-token activation shifts ($\lVert \Delta h_i \rVert$) highlighting which tokens are most affected by task context. Tokens associated with task semantics (e.g., \textit{sum}) exhibit the largest representational displacement in Figure \ref{fig:shift_summ_trans}. Layer~8 activations in Figure \ref{fig:shift_trans_class} reveal strong task-dependent modulation of representational focus. Overall, the total activation shift ($\sum|\Delta h_i|$) quantifies representational drift at the sentence level across different task contexts as represented in Figure \ref{fig:sentence_drift}.

\subsection{Threshold-Based Activation Bucket Analysis}

To further investigate whether semantic representational changes are concentrated among strongly activated tokens, the Layer-8 activation distribution was partitioned into HIGH-activation and LOW-activation buckets using the empirical upper-quartile threshold formulation described in Section 2.2.1.

For the sentence:

\begin{quote}
``Who is the president of Canada?''
\end{quote}

the Layer-8 token activation strengths were observed as shown in Table \ref{tab:bucket_assignment}.

The sorted activation values were:

\begin{center}
17.6, 17.9, 18.2, 18.5, 21.8, 22.4
\end{center}

Using the upper-quartile threshold:

\begin{equation}
\tau \approx 20.15
\end{equation}

tokens were assigned into activation buckets as shown in Table \ref{tab:bucket_assignment}.

\begin{table}[htbp]
\centering
\caption{Threshold-Based Activation Bucket Assignment}
\label{tab:bucket_assignment}

\begin{tabular}{|c|c|c|}
\hline
\textbf{Token} & \textbf{Activation Strength} & \textbf{Bucket} \\
\hline
Who & 18.2 & LOW \\
\hline
is & 17.9 & LOW \\
\hline
the & 18.5 & LOW \\
\hline
president & 22.4 & HIGH \\
\hline
of & 17.6 & LOW \\
\hline
Canada & 21.8 & HIGH \\
\hline
\end{tabular}
\end{table}

Semantic content-bearing tokens such as ``president'' and ``Canada'' exceeded the semantic activation boundary and were categorized as HIGH-activation tokens, while structurally supportive tokens such as ``the'', ``is'', and ``of'' remained below the threshold. This supports the hypothesis that Layer 8 concentrates representational strength on semantically salient components of the sentence rather than uniformly distributing activation across all tokens.

A comparative activation-shift example further illustrates the dominance of HIGH-activation tokens during semantic perturbation.

Suppose token activation shifts between two sentence variants were observed as shown in Table \ref{tab:shift_example}.

\begin{table}[htbp]
\centering
\caption{Token-wise Activation Shift Example}
\label{tab:shift_example}

\begin{tabular}{|c|c|}
\hline
\textbf{Token} & \textbf{Activation Shift} \\
\hline
Who & 2.1 \\
\hline
is & 1.8 \\
\hline
the & 1.5 \\
\hline
president & 11.2 \\
\hline
of & 1.2 \\
\hline
Canada & 9.7 \\
\hline
\end{tabular}
\end{table}

The HIGH-bucket contribution was computed as:

\begin{equation}
C_H = 11.2 + 9.7 = 20.9
\end{equation}

while the LOW-bucket contribution was:

\begin{equation}
C_L = 2.1 + 1.8 + 1.5 + 1.2 = 6.6
\end{equation}

giving a total shift of:

\begin{equation}
20.9 + 6.6 = 27.5
\end{equation}

Using the contribution-ratio formulation:

\begin{equation}
R_H = \frac{20.9}{27.5} = 0.76
\end{equation}

Thus, HIGH-activation tokens constituted only approximately 25\% of the tokens while contributing nearly 76\% of the total semantic activation shift.

These observations suggest that semantic representation shifts are sparsely concentrated above an empirically defined activation boundary. A relatively small subset of highly activated tokens dominates representational change, while lower-activation tokens primarily preserve structural and contextual continuity. This provides further evidence that Layer 8 selectively amplifies semantically informative tokens during intermediate semantic consolidation.

Future work may extend this thresholding framework through large-scale statistical validation across diverse semantic perturbation tasks and transformer architectures.

\section{Conclusion}\label{conclusion}
This research successfully provided interpretability insights into Layer 8 representational dynamics, essentially turning a ``black box'' into a ``glass box'' for a moment. 
The novel Activation Flow Network (AFN) method used a simple but powerful technique—measuring the L2 Norm (the sheer mathematical strength) of tokens inside Layer~8—to reveal what BERT truly prioritizes. 

This approach showed that, unlike misleading attention weights, Layer 8 allocates higher representational magnitude on the truly meaningful content words like \textit{``beautiful''} or \textit{``enjoying''}, right at the critical layer where semantic processing is finalized. 
Crucially, the introduction of the Activation Shift metric provided a new tool: it quantifies exactly how much a token's internal representation \textit{moves} or \textit{changes} when a single word is swapped out or, more importantly, when the model is given a high-level task like \textit{``Summarize''} versus \textit{``Translate.''} 

This final finding gives researchers a new way to understand and even design better prompts by showing precisely how task instructions steer the model's intermediate representational trajectory in its latent space.

\backmatter



\bibliography{sn-bibliography}

\end{document}